\documentclass[10pt,twocolumn,letterpaper]{article}

\usepackage{cvpr}
\usepackage{times}
\usepackage{epsfig}
\usepackage{graphicx}
\usepackage{amsmath}
\usepackage{amssymb}
\usepackage{comment}
\usepackage{multirow} 
\usepackage{array}  

\usepackage{color}
\usepackage{verbatim}
\usepackage{booktabs}   
\usepackage{bm}
\usepackage{float}

\usepackage[breaklinks=true,bookmarks=false]{hyperref}

\cvprfinalcopy % *** Uncomment this line for the final submission

\def\cvprPaperID{****} % *** Enter the CVPR Paper ID here

\begin{document}

%%%%%%%%% TITLE
\title{Unsupervised Facial Action Unit Intensity Estimation via Differentiable Optimization}

\author{Xinhui Song\\
NetEase Fuxi AI Lab\\
{\tt\small songxinhui@corp.netease.com}
\and
Tianyang Shi\\
NetEase Fuxi AI Lab\\
{\tt\small shitianyang@corp.netease.com}
\and
Tianjia Shao\\
Zhejiang University\\
{\tt\small tianjiashao@gmail.com}
\and
Yi Yuan\\
NetEase Fuxi AI Lab\\
{\tt\small yuanyi@corp.netease.com}
\and
Zunlei Feng\\
Zhejiang University\\
{\tt\small zunleifeng@zju.edu.cn}
\and
Changjie Fan\\
NetEase Fuxi AI Lab\\
{\tt\small fanchangjie@corp.netease.com}
}

\maketitle
\begin{figure*}[htbp]
	\begin{center}
		\includegraphics[width=\linewidth]{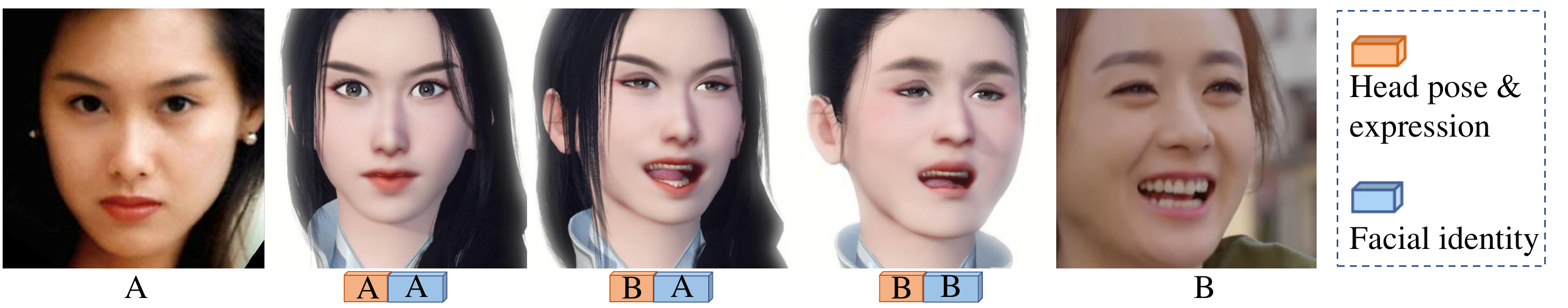}
	\end{center}
	\caption{An application of our method is expression transfer. We first estimate the physically meaningful facial parameters (i.e., head pose, AU parameters and identity parameters) from the input image A/B with our differentiable optimization framework GE-Net.  Then we transfer the pose and expression from B to A while keeping A's identity by recombining the facial parameters. The middle images are rendered using the Justice Face model with the estimated facial parameters.}
	\label{overview}
\end{figure*}

%%%%%%%%% ABSTRACT
\begin{abstract}
  The automatic intensity estimation of facial action units (AUs) from a single image plays a vital role in facial analysis systems. 
  One big challenge for data-driven AU intensity estimation is the lack of sufficient AU label data.
  Due to the fact that AU annotation requires strong domain expertise, it is expensive to construct an extensive database to learn deep models. 
  The limited number of labeled AUs as well as identity differences and pose variations further increases the estimation difficulties. 
  Considering all these difficulties, we propose an unsupervised framework GE-Net for facial AU intensity estimation from a single image, without requiring any annotated AU data. Our framework performs \emph{differentiable optimization}, which iteratively updates the facial parameters (i.e., head pose, AU parameters and identity parameters) to match the input image. GE-Net consists of two modules: a generator and a feature extractor. The generator learns to ``render" a face image from a set of facial parameters in a differentiable way, and the feature extractor extracts deep features for measuring the similarity of the rendered image and input real image. After the two modules are trained and fixed, the framework searches optimal facial parameters by minimizing the differences of the extracted features between the rendered image and the input image. 
  Experimental results demonstrate that our method can achieve state-of-the-art results compared with existing methods.
\end{abstract}

\section{Introduction}

Facial expression analysis is of great interest to many researchers in computer vision, computer graphics, and psychology. 
Automated facial expression analysis from a single image enables numerous applications such as human-robot interaction and behavioral and psychological research.
The Facial Action Unit (AU) intensity estimation \cite{friesen1978facial} is one of the main building blocks in single-image facial expression analysis, which aims to estimate anatomically meaningful parameters (i.e., muscle movements) instead of PCA parameters of a 3D morphable model~\cite{blanz1999morphable}.
The seminal work of Ekman and Friesen~\cite{friesen1978facial} develops a Facial Action Coding System (FACS) for describing facial expressions, which are produced by the movements of facial muscles under the skin. 
Nearly any anatomically possible facial expression can be coded by a combination of AUs.

Existing AU intensity estimation methods are broadly divided into fully supervised methods \cite{baltruvsaitis2015cross,li2017action,zhou2017pose,batista2017aumpnet} and weakly supervised methods \cite{zhao2016facial,zhang2018weakly,zhang2018bilateral}. 
Training a fully supervised model requires a large set of labeled samples. However, AU annotation requires strong domain expertise, resulting in very high time-and-labor costs to construct a large database.
Due to this demanding labeling process, datasets in the literatures (\eg, CK+ \cite{lucey2010extended}, MMI \cite{pantic2005web},  AM-FED \cite{mcduff2013affectiva}, DISFA \cite{mavadati2013disfa}, BP4D \cite{zhang2013high}) restrict the number of coded AUs, samples, and subjects. 
The weakly supervised methods focus on various types of human knowledge with limited annotation datasets. 
Nevertheless, they still require thousands of AU annotations, and the human defined knowledge restricts the space of anatomically plausible AUs.
Besides, there are many challenges in automatic AU intensity estimation so that the performance of AU intensity estimation is often unsatisfactory. 
These challenges include large variations in pose, identity, illumination, and occlusion in the unconstrained environment. 
Among them, pose and identity have always been essential challenges because they can dramatically increase variances of face images of different people even when they have the same expression.

%Existing methods always leverage one of them and there is no general framework to incorporate different types of knowledge. 

%介绍我们手头可以有什么样的数据
%\tianjia{Dont use Generating Segmentation Network, espeically do not use segmentation. It is very misleading.}

Considering the above challenges, we propose a framework for unsupervised estimation of facial action unit intensity from a single image, which can jointly estimate the facial parameters (i.e., head pose, AU parameters and identity parameters) without annotated AU data. 
The core of our framework is \emph{differentiable optimization}, which iteratively updates the facial parameters to match the input image.
It is achieved with the help of a novel network architecture called GE-Net. GE-Net consists of two modules: a generator which learns to ``render" a face image from a set of facial parameters in a differentiable way, and a feature extractor which extracts deep features for measuring the similarity of the rendered image and input real image. 
After the two modules are trained, we integrate them into our framework, and the facial parameters can be fitted by minimizing the loss of the extracted features between the rendered image and the input real one. 

Specifically, our generator is trained based on a 3D parametric face model called Justice Face from an online game \emph{Justice}\footnote{https://n.163.com/}.
The model parameters have explicit physical meanings, including AU parameters and identity parameters.
The AU parameters can express all the AUs from FACS, and the identity parameters describe identity-specific face shapes by controlling the attributes of facial components, including position, orientation, and scale, as introduced in the work \cite{F2p}.
We adopt a face semantic segmentation network as our feature extractor, whose intermediate feature maps focus more on the facial shape contents rather than the raw appearance. Hence it can effectively reduce the large domain gap between the rendered image and the real one (see Sec. \ref{ablation_sec} for details).

%\xinhui{ While F2p \cite{F2p} only learns identity parameters to the rendered image by a simply generator.}

Our method can be used in many applications, one of which is expression transfer. We can exchange the AU parameters of different individuals, which is shown in Fig. \ref{overview}. 
With image A or B as input, our framework can predict their facial parameters. After re-combining the head pose, AU parameters, and identity parameters, we can transfer the pose and expression from B to A while keeping A's identity. 

%Similarly, the AU parameters can be transferred to other 3D facial models, which is constructed according to FACS.

Our main contributions are summarized as the following: To the best of our knowledge, we firstly propose a deep learning framework for unsupervised estimating the facial AU intensity from a single image, without requiring annotated AU data. 
What's more, the novel framework enables differentiable optimization, which consists of a generator performing differentiable rendering and a feature extractor measuring similarity of the rendered image and real image.
%It performs differentiable optimization to fit the facial parameters to the input image. 
Large experiments show that the proposed method achieves state-of-the-art results on the AU intensity estimation. Meanwhile, our method can activate more AUs than existing AU annotated databases, such as jaw left and upper lip raiser.
%The proposed GE-Net achieves the state-of-the-art results on the AU intensity estimation, and we can activate more AUs than existing AU annotated databases, such as jaw left and upper lip raiser. 

\section{Related Work}
%AU estimation has received great attention over the past years.
%In the following paragraphs, we review the literatures most relative to ours.

\subsection{Face Reconstruction Methods based on 3DMM}
In the past years, many methods \cite{zhu2015high,zhu2016face,guo2018cnn,zhou2019dense} describe a 3D facial shape using the parameters of the 3DMM \cite{blanz1999morphable}, which are constructed with PCA. 
Early methods \cite{hpen,cao2014displaced} use a regressor to directly regress the facial landmark locations from a single image for fitting 3DMM parameters.
Recently, the approaches of regressing 3DMM parameters using CNNs and fitting 3DMM to the 2D images have become popular. 
Jourabloo \etal ~\cite{jourabloo2016large} use cascade CNN regressors to regress the 3DMM parameters (identity, expression and pose parameters). 
Bindita \etal ~\cite{chaudhuri2019joint} present a single end-to-end network to jointly predict the bounding box locations and 3DMM parameters for multiple faces. Zhu \etal ~\cite{zhu2016face,zhu2017face} perform multiple iterations of a single CNN to fit the 3DMM parameters. 
Different from these 3DMM-based methods, our goal is to predict the facial AU intensity with explicit anatomical meanings, while the 3DMM parameters lack such physical meanings.

% 3DMM parameters regressed by the previous methods are not interpretable. We first emphasize the distinction between the methods based on 3DMM, yet different goals of expression regression. The former regress the pca parameters. The latter regress the interpretable parameters, which are more difficult task. 

% In addition, while the approach of regressing 3DMM parameters need large of annotated datasets, fitting 3DMM to the 2D images only focus geometric details. Our method can fit the interpretable parameters from a portrait in different views by the deep features, and it does not require any AU annotation.  

\begin{figure*}
	\begin{center}
		\includegraphics[width=\linewidth]{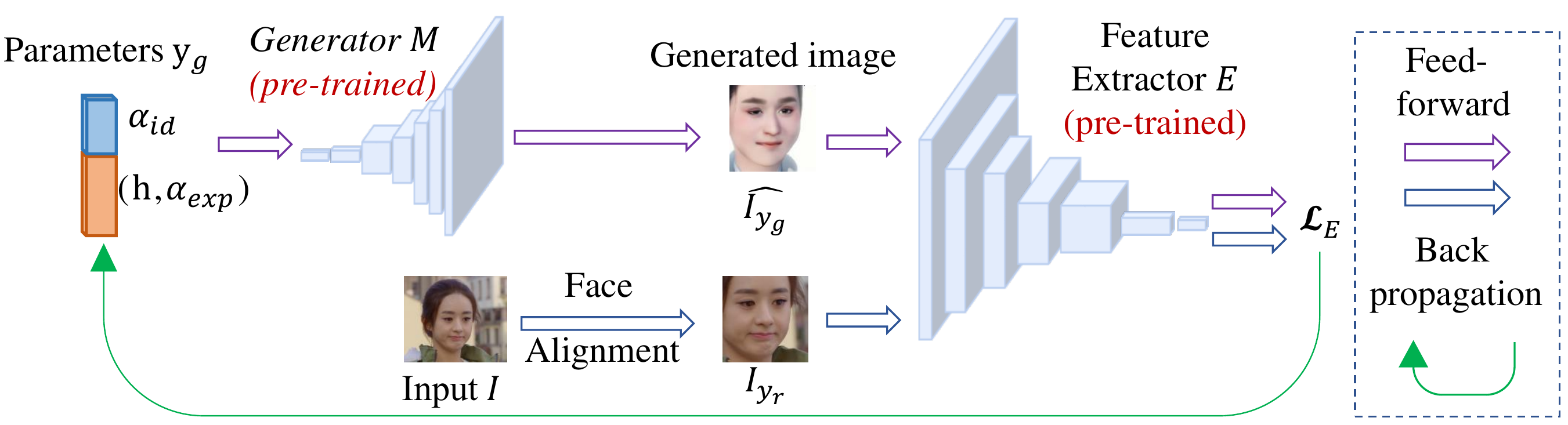}
	\end{center}
	\caption{The pipeline of the proposed method. A generator and a feature extractor are trained in the training phase. In the testing stage, we fix the parameters of the generator and the feature extractor. The framework shows the process that updates the facial parameters by minimizing the differences between the real image and a generated image. Empirically, the optimal facial parameters can be obtained after 30 iterations for an image.}
	\label{network}
\end{figure*}

\subsection{ AU intensity Estimation Methods}
%TODO 相对于au 估计， AU预测的方法是做分类的任务。

Similar to AU detection \cite{niu2019local},
some methods \cite{kaltwang2012continuous,kaltwang2015latent,kaltwang2015doubly,rudovic2014context} use traditional machine learning frameworks to estimate AU intensity. 
Recently, some works apply CNNs for AU intensity estimation. 
Tadas \etal ~\cite{baltruvsaitis2015cross} present a facial action unit intensity estimation system based on geometry features (shape parameters and landmark locations) and appearance (Histograms of Oriented Gradients). 
The median-based feature normalization is used to account for a person-specific neutral expression. 
Li \etal ~\cite{li2017action,li2017eac} design different CNNs to regress AU parameters, where the CNNs focus on different facial regions independently. 
Batista \etal ~\cite{batista2017aumpnet} and Zhou \etal ~\cite{zhou2017pose} apply DNNs to estimate AU intensity under multiple head poses. 
However, these supervised methods require a large number of training samples, while AU annotation needs strong domain expertise, and it requires great efforts to construct such extensive database. 
Besides, limited by the number of annotated samples, above methods usually suffer from an over-fitting problem. 
%On the contrary, our method can estimate more AUs without any AU annotation. 
Due to the lack of AU annotation datasets, several works \cite{zhao2018learning,zhang2018weakly,zhang2018bilateral} use the weakly supervised training paradigm to train deep models with incomplete or inaccurate annotations, where prior knowledge and domain knowledge are involved. 
However, these methods still require thousands of annotations, and they add prior knowledge, which restricts the space of anatomically plausible AUs. 
In contrast, our method can estimate more AUs than existing AU annotated databases, without requiring any annotated data.

\subsection{Neural Render} 
Recent works on differentiable rendering achieve differentiability in various ways. Kato \etal ~\cite{kato2018neural} achieve differentiability by an approximate gradient for the rasterization operation. 
GQN \cite{eslami2018neural} shows a powerful neural renderer, which learns to represent synthetic scenes using the input images of a scene taken from different viewpoints, but the geometry of objects in the scene is simple. 
MOFA \cite{tewari2017mofa} learns a decoder using a neural renderer model, which is an expert-designed generative model. 
The expert-designed generative model is difficult to describe the complicated rendering process with the illumination condition and complex shading explicitly. Nguyen-Phuoc \etal ~\cite{nguyen2018rendernet} propose a CNN architecture to render 3D objects from a 3D voxel grid. 
Shi \etal ~\cite{F2p} learn a generator as a neural render with fixed camera parameters. 
Inspired by \cite{F2p}, we extend the generator to imitate the face rendering process from meaningful facial parameters including identity, expression and head pose.

\section{Method}
Our goal is to estimate the intensity of facial AUs as well as the face pose and identity from a single image via \emph{differentiable optimization}, without requiring any labeled data.   
The optimization framework GE-Net consists of a generator $\mathcal{M}$ and a feature extractor $\mathcal{E}$. 
The generator $\mathcal{M}$ imitates the rendering process from the facial parameters $y_g$ (i.e., head pose $h$, AU parameters $\alpha_{exp}$ and identity parameters $\alpha_{id}$) to the rendered image $I_{y_g}$, while the feature extractor $\mathcal{E}$ extracts deep features from images, which are used to measure the similarity between the real input image and the rendered image for fitting the meaningful parameters.
The generator is trained with the help of a 3D parametric face model called Justice Face from the game \emph{Justice}. The model parameters have explicit physical meanings, including AU parameters $\alpha_{exp}$ and identity parameters $\alpha_{id}$.
A complete process of our method is summarized as follows:\\
\emph{Stage 1.} Train the generator $\mathcal{M}$ and the feature extractor $\mathcal{E}$.\\
\emph{Stage 2.} Fix the parameters of $\mathcal{M}$ and $\mathcal{E}$, and iteratively update the facial parameters $(h,\alpha_{exp},\alpha_{id})$ according to the loss between the input image and the generated one.
Empirically, the facial parameters can closely describe the input image after 30 iterations.

\begin{figure*}[t]
	\begin{center}
		\includegraphics[width=\linewidth]{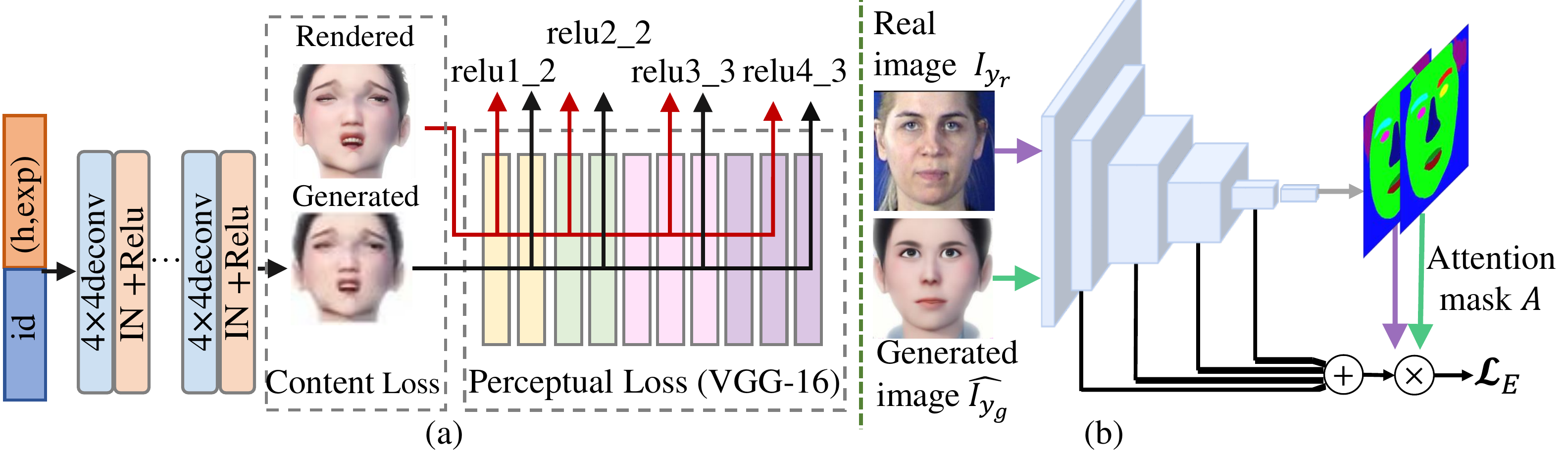}
	\end{center}
	\caption{The generator $\mathcal{M}$ (a) and the feature extractor $\mathcal{E}$ (b). The generator learns to map a set of meaningful facial parameters to a rendered image by combining the content loss and perceptual loss. IN means instance normalization \cite{ulyanov2016instance}. A face segmentation network is trained as the feature extractor. The similarity between the input real image and the generated image is measured via the $L_1$ loss of the feature maps pixel-wisely multiplied with the attention masks. The output probability map of every local region serves as an attention mask for one of the feature maps.}
	\label{generator}
\end{figure*}

\subsection{3D Face Model}
% 描述一下我们的3D模型
%We use a 3D facial model by bone-driven, and then the identity parameters are baked from the mesh with the same topology.  
The parametric face model Justice Face contains the identity parameters and AU parameters. The identity parameters $\alpha_{id}$ control the attributes of facial components to create a neutral face for a specific identity, including the position, orientation, and scale as recommended by the method \cite{F2p}. 
%The offsets relative to the neutral face are baked to AU parameters, which can drive the facial muscle movement. 
The AU parameters $\alpha_{exp}$ control the expression by driving the facial muscle movements relative to the neutral face. Specifically, these AU parameters include eyes closed, upper lid raiser, cheek raiser, inner brows raiser, outer brows raiser, brow lower, jaw open, nose wrinkle, upper lip raiser, down lip downer, lip corner pull, mouth press, lip pucker, upper lips close, lower lips close, cheek puff, lip corner depressor, jaw left and jaw right. 
The corresponding maps between AUs and our AU parameters are shown in Table 2 of the appendix. 
We can adjust the AU and identity parameters to generate an identity-specific expression, which is formulated below:  

% (The corresponding map between AUs and our AU parameters please refer to the supplement material). 

\begin{equation}
S = \overline{S} + \alpha_{id}A_{id} + \alpha_{exp}A_{exp},
\end{equation}
where $S$ is a synthesized 3D face and $\overline{S}$ is the base face. $A_{id}$ is the facial component offsets to the base face in position, orientation, and scale, and the identity parameters $\alpha_{id}$ are the weights of these offsets. $A_{exp}$ is the offsets of facial muscle movements relative to the neutral face. The AU parameters $\alpha_{exp}$ are the corresponding weights and represent the facial AU intensity. 
%The facial model comes from the game ``Justice", in which the parameters $(\alpha_{id}, \alpha_{exp})$ are interpretable.

\subsection{Generator}
The generator $\mathcal{M}: (y_{g}) \longrightarrow{} I_{y_g}$ learns to map a set of meaningful facial parameters $y_{g}=(h,\alpha_{exp},\alpha_{id})$ to a rendered image $I_{y_g}$ based on the Justice Face. $y_{g}$ is a 269-dimensional vector with head pose $h\in \mathbb R^{1\times2}$, AU parameters $\alpha_{exp} \in R^{1\times23}$ and identity parameters $\alpha_{id} \in \mathbb R^{1\times244}$. 
The ground truth $I_{y_g} \in \mathbb R^{H\times W\times 3}$ is the corresponding rendered image from the Justice Face, and the predicted one is marked as the generated image $\hat{I_{y_g}}$. 
Our generator $\mathcal{M}(y_g)$ consists of eight transposed convolution layers, which is similar to the network of DCGAN \cite{radford2015unsupervised}. 
We use instance normalization \cite{ulyanov2016instance} instead of batch normalization \cite{ioffe2015batch} in the generator for improving training stability. 
The generator is differentiable so that the facial parameters can be updated by gradient descent.

Fig. \ref{generator}(a) shows the architecture of our generator. 
To make the generated images indistinguishable from the rendered images, we adopt a content loss $\mathcal{L}_{app}$ and a perceptual loss $\mathcal{L}_{per}$~\cite{johnson2016perceptual}  between the rendered image $I_{y_g}$ and the predicted one $\hat{I_{y_g}}$. The content loss  $\mathcal{L}_{app}(y_g)$ penalizes large changes in the raw pixel space, and we use $l_1$ loss function rather than $l_2$ for $\mathcal{L}_{app}$ so as to encourage less blurring. The content loss is defined as:
\begin{equation}
\begin{aligned}
\mathcal{L}_{app}(y_g) &= E_{y_g\sim u(y_g)}[\Arrowvert I_{y_g}-\hat{I_{y_g}}\Arrowvert_{1}],
\end{aligned}
\label{l1loss}
\end{equation}
%  &= E_{y_g\sim u(y_g)}[\Arrowvert I_{y_g}- \mathcal{M}(y_g)\Arrowvert_{1}].
where $\hat{I_{y_g}} = \mathcal{M}(y_g)$ and $u(y_g)$ means the uniform distribution. The input parameters $y_g$ are sampled from a multidimensional joint uniform distribution $u(y_g)$.

The perceptual loss $\mathcal{L}_{per}(y_g)$ is added to explicitly constrain the generated result to have the same content information as the rendered one, and it is defined as: 
\begin{equation}
\begin{aligned}
\mathcal{L}_{per}(y_g) &= E_{y_g\sim u(y_g)}[\Arrowvert \mathcal{F}(\hat{I_{y_g}})-\mathcal{F}(I_{y_g})\Arrowvert_{2}],
\end{aligned}
\label{eq_perceputalloos}
\end{equation}
where $\mathcal{F}$ denotes the \text{relu2\_2, relu3\_3, relu4\_3} feature maps in VGG16 \cite{simonyan2014very}, which is pre-trained on the image recognition task. 
In this way, we can train the generator $\mathcal{M}$ by optimizing the following loss function: 

\begin{equation} 
\mathcal{L}_{\mathcal{M}}(y_g) =\mathcal{L}_{app}(y_g) +\lambda\mathcal{L}_{per}(y_g),
\end{equation}
where $\lambda$ balances the multiple objectives.

During training, all the rendered images are aligned via five facial landmarks using affine transformation, including left eye, right eye, nose, left mouth corner and right mouth corner, which are obtained by OpenFace \cite{baltrusaitis2013constrained,zadeh2017convolutional}.
The convolution kernel size in the generator is $4\times4$, and the stride of each transposed convolution layer is 2. 
The Instance-Normalization and ReLU activation are embedded in $\mathcal{M}$ after every transpose convolution layer, except for its output layers. $\mathcal{M}$ is trained with mini-batch stochastic gradient descent (SGD) and the mini-batch size is 64.
All the weights are initialized from a zero-centered normal distribution with a standard deviation of 0.02. 
The learning rate is 0.001, and the learning rate decay is set to 1\% per three epochs. After training of 500 epochs, the generator successfully learns how to map each facial parameter to its corresponding component/muscle movement and ``render" a visually plausible image.
%The goal of training the $\mathcal{M}$ is to split the facial identity parameters and facial AU parameters. 
%Each of the parameters can drive a meaningful physical offset as the renderer. \tianjia{(dont understand the goal.)} \xinhui{ The generator is trained for 500 epochs so that each of the facial parameters can drive the corresponding facial components.}

% Finally, we aim at minimizing the $\mathcal{L}_{\mathcal{M}}(y_g)$.
%	We randomly generate the parameters by a unified distribution so that $\mathcal{M}$ can flexibly translate the parameters. 

\subsection{Feature Extractor}
As there is a large domain gap between the generated image and the input real image, to effectively measure their similarity, a face segmentation network is trained as a feature extractor $\mathcal{E}$ in Fig. \ref{generator}(b). The similarity is computed on the feature maps of the first four layers of $\mathcal{E}$. Specifically, an attention mechanism \cite{cho2014learning} is deployed in the feature extractor. 
It allows the feature maps to attend to different parts of the face, which means the feature maps have different influences on the loss. 
The output probability maps of the facial segmentation network serve as the attention masks, which are represented as the pixel-wise weights $A$ in Eqn. (\ref{featExtract}). 
Specifically, the probability map of every facial local region is utilized as an attention mask of each feature map. For example, the feature map of the first layer is sensitive to the eyes, and 
the eyes' probability map $A$ is element-wisely multiplied to the feature map of the first layer. 
Suppose $C$ represents the feature map, and the extracted features can be defined as:
\begin{comment}
\begin{figure}
\begin{center}
%\fbox{\rule{0pt}{2in} \rule{1\linewidth}{0pt}}
\includegraphics[width=\linewidth]{images/feature-extractor.jpg}
\end{center}
\caption{The feature extractor $\mathcal{E}(x)$. The semantic segmentation network is used to extract the face feature. The similarity is measured via the $L_1$ loss of the feature maps pixel-wise multiply attention mask between the real image and the generated image. The output probability map of every local region is as an attention mask for one of the feature maps.}
\label{featExtractor}
\end{figure}
\end{comment}
\begin{equation}
\label{featExtract}
\mathcal{E}(I) = \sum_{i=0}^{N}  \beta_{i} A_i(I) C_i(I),
\end{equation}
where $I$ is the input of the feature extractor. $N$ feature maps are extracted, and $\beta$ represents the weight of each feature map $C_i$. We set different attention masks $A_i$ for the feature maps by the probability maps of the facial segmentation result.

The architecture of the feature extractor network is based on Resnet-50 \cite{he2016deep}. 
We set the stride [2, 1, 1, 1] at four sub-layers, remove the fully connected layers and change the average pooling layer to a $1\times 1$ convolution layer.  
%After the $1\times 1$ convolution,
The output is 1/8 size of the input. 
We further perform up-sampling with factor 8 to get an upsampled output with the same size as the input. 
The network is pre-trained on the ImageNet \cite{deng2009imagenet} and is fine-tuned with the Helen Dataset \cite{le2012interactive} with the pixel-wise cross-entropy loss. The learning rate is set to 0.001.

\begin{table*}[ht]
	\caption{The Intra-Class Correlation (ICC) and Mean Absolute Error (MAE) on FERA2015 and DISFA. Bold numbers indicate the best performance. Underline numbers indicate the second best. (*) indicates results taken from the reference.} 
	\label{compare_ICC_MAE}
	\begin{center} 
		\scalebox{0.9}{
			\begin{tabular}{|p{3mm}<{\centering}|p{18mm}<{\centering}|p{4mm}<{\centering}p{4mm}<{\centering}p{4mm}<{\centering}p{5mm}<{\centering}p{4mm}<{\centering}p{5mm}<{\centering}|p{4mm}<{\centering}p{4mm}<{\centering}p{4mm}<{\centering}p{4mm}<{\centering}p{4mm}<{\centering}p{5mm}<{\centering}p{4mm}<{\centering}p{4mm}<{\centering}p{4mm}<{\centering}p{4mm}<{\centering}p{4mm}<{\centering}p{4mm}<{\centering}p{5mm}<{\centering}|} %12 au 
				\hline
				~& Database & \multicolumn{6}{c|}{FERA2015}& \multicolumn{13}{c|}{DISFA} \\
				\cline{2-21}
				~ & AU & 06 & 10 & 12& 14& 17 & Avg & 01 & 02& 04 &05& 06& 09& 12& 15 & 17& 20& 25 &26 & Avg \\
				\hline
				\multirow{6}{*}{\rotatebox{90}{ICC}} & KBSS \cite{zhang2018weakly}* & .76 & .73 & .84 & .45 & .45 & .65         & .23 & .11 & .48 & .25 & .50 & .25 & .71 & .22 &.25 & .06& \underline{.83} &  .41 & .36 \\
				~ & KJRE \cite{zhang2019joint}* &.71 &.61 & {\bf.87} &.39 & .42 & .60               & .27 & .35 & .25 & .33 & .51 & .31 & .67 & .14 & .17 & .20 & .74 & .25 & .35  \\
				~ & LBA \cite{haeusser2017learning}* & .71 & .64 & .81 & .23 & .50  & .58  & .04 & .06 & .39 & .01 & .41 & .12 & \underline{.73} & .13 & .27& .10 & .82 & .43 & .29 \\
				~ & 2DC \cite{linh2017deepcoder}* &.76 & .71 & \underline{.85} & .45 & .53  & .66     & {\bf.70} & .55 & \underline{.69} & .05 & \underline{.59} & \underline{.57} & {\bf.88} & .32 & .10 & .08 & {\bf.90} & .50 &  .50\\
				~ & CFLF \cite{zhang2019context}*& .77 & .70 & .83 & .41 & {\bf.60}  & .66 & .26 & .19 & .46 & .35 & .52 & .36 & .71 & .18 & .34 & .21 & .81 &.51 & .41\\
				%~ & OPENFACE \cite{baltruvsaitis2015cross} & & & & &  & &.59 & \underline{.58} & .40 & \underline{.41} & .07 & .37 & .55 & .27 & .44 & .32 &.45 & .25 & .39 \\
				~ &  GENet-O  & \underline{.67} & \underline{.80} & .71 & \underline{.61} & .50  & \underline{.66} & .51 & \underline{.63} & .58 & .65 & .53 & .62 & .48 & \underline{.55} & \underline{.50} & \underline{.37} & .70 &\underline{.63} & \underline{.59} \\
				~ &  GE-Net  & {\bf.69} & {\bf.85} & .73 & {\bf.63} & \underline{.53}  & {\bf.68} & \underline{.66} & {\bf.67} & {\bf.73} & {\bf.71} & {\bf.60} & {\bf.59} & .60 & {\bf.66} & {\bf.58} & {\bf.45} &.80 &{\bf.70} & {\bf.64} \\
				\hline

				\multirow{6}{*}{\rotatebox{90}{MAE}} & KBSS \cite{zhang2018weakly}*& .74 & .77 & .69 & .99 & .90  & .82  & .48 & .49 &  .57 & {\bf.08} & {\bf.26} & \underline{.22} & \underline{.33} & \underline{.15} & .44 & \underline{.22} & .43 & \underline{.36} & .336  \\	
				~ & KJRE \cite{zhang2019joint}* & .82 & .95 & .64 & 1.08 & .85  &.87      & 1.0 & .92 & 1.9 & .70 & .79 & .87 & .77 & .60 & .80  & .72 & .96 & .94 & .91  \\	
				~ &LBA \cite{haeusser2017learning}* & .64 & .80 & .56 & 1.10 & .62 & .74  &.43&  .29 &.51 &\underline{.10} &\underline{.30} &{\bf.19} &{\bf.30}& {\bf.11}& \underline{.31}& {\bf.14} &.40 & .38& {\bf.29} \\			
				
				~ & 2DC \cite{linh2017deepcoder}* & .87 & .84 & .92 & .67 & .73  & .81 & .57 & .62 & .73 & .51 & .66 & .55 & .50 & .52 & .78 & .42 & .61 & .74 & .61\\
				
				~ & CFLF \cite{zhang2019context}* & .62 & .83 & .62 & 1.00 & .63  &.74      & {\bf.33} & {\bf.28} & .61 & .13 & .35 & .28 & .43 &.18 & {\bf.29} & .16 & .53 & .40 & .329 \\
				
				%~ & OPENFACE \cite{baltruvsaitis2015cross} &  & &  &  &   & & .40 & .43 &.44 &.20 & .52 & .51& .43 & .25 &.45 &.47& .50 & .44 & .42 \\
				~ &  GENet-O & \underline{.49} & \underline{.72} & \underline{.45} & \underline{.43} & \underline{.34}  &\underline{.49} & .39 & .35& {\bf.37} & .17 & .34 &.39& .55& .19 & .43 &.58& {\bf.22} & .32 & .36 \\ 
				~ &  GE-Net & {\bf.45} & {\bf.60} & {\bf.37} &{\bf.40} & {\bf.27}  & {\bf.42}  & \underline{.36} & \underline{.33} & \underline{.38} & .14 &.34 &.36 & .57 & .18 &.42 & .26& \underline{.27} & {\bf.32} &\underline{.329}\\
				\hline
				
			\end{tabular}
		}
	\end{center}
\end{table*}

\subsection{Fitting}
After the generator and the feature extractor are learned, we fix their parameters and iteratively optimize the facial parameters (i.e., head pose, AU parameters and identity parameters) to match the input real image. 
%The facial parameter regressor can be modeled linearly in representation space. \tianjia{(what do you mean? dont understand. no regressor mentioned before.)} 
To fit the facial parameters to a real face image with expression, the generator $\mathcal{M}$ and the feature extractor $\mathcal{E}$ serve as a bridge. 
We optimize the facial parameters by measuring the similarity of the extracted features from $\mathcal{E}$ between the input real image and the generated image from $\mathcal{M}$. The extracted features can teach the generator to produce faces that match firmly with the real image in the feature space.

Fig. \ref{network} shows the pipeline of our proposed method. The meaningful facial parameters can be updated using gradients because the renderer is differentiable.
We update the facial parameters by back-propagating the loss on extracted features, which explicitly constrain the ``rendered" result to have similar facial shape contents as the real image. The loss is defined as:
\begin{comment}
\begin{equation}
\mathcal{L}_{E}(y_g,I_{y_r}) = \Arrowvert \mathcal{W}(\mathcal{M}(y_g))\mathcal{E}(\mathcal{M}(y_g)) - \mathcal{W}(I_{y_r})\mathcal{E}(I_{y_r})\Arrowvert_{1},
\label{featExtract}
\end{equation}
\end{comment}
\begin{equation}
\mathcal{L}_{E}(y_g,I_{y_r}) = \Arrowvert \mathcal{E}(\mathcal{M}(y_g)) - \mathcal{E}(I_{y_r})\Arrowvert_{1}, 
\end{equation}
where $I_{y_r}$ is the input real image aligned with the base face using five facial landmarks.
%The optimization is solved with the gradient descent method. 
The parameters are updated with the gradient descent method by:

\begin{equation}
y_g \leftarrow  y_g - l_r\frac{\partial \mathcal{L}_{E}}{\partial y_g},
\end{equation}
where $l_r$ represents the learning rate. The range of $y_g$ is clipped to [0,1].

\begin{figure}[t]
	\begin{center}
		\includegraphics[width=0.99\linewidth]{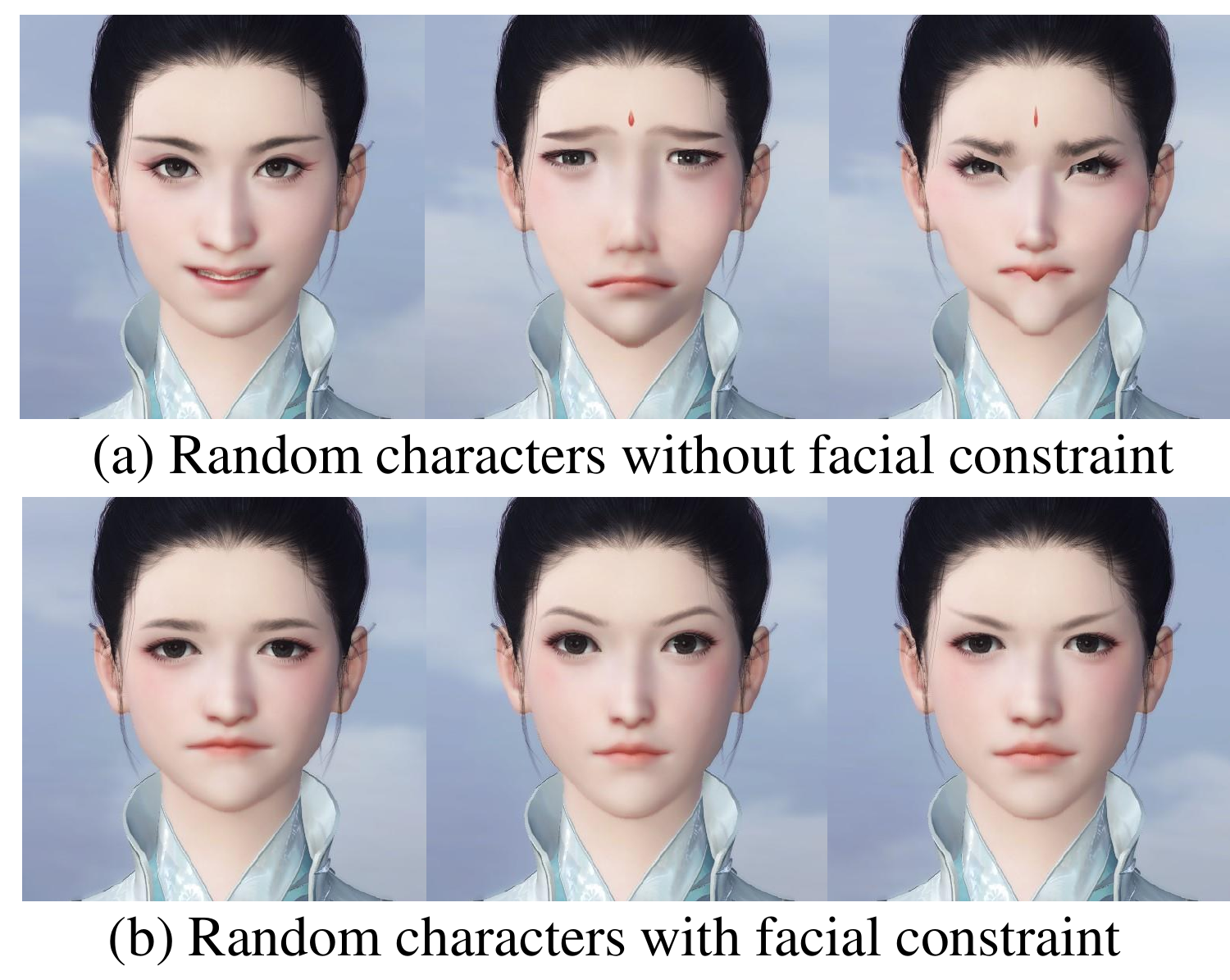}
	\end{center}
	\caption{ Randomly generated faces without and with the facial constraint. The facial constraint can restrict the space of the identity parameters and generate reasonable and neutral faces. }
	\label{generate_images}
\end{figure}

However, the identity parameters can describe a large-scale 3D space, in which some points don't exist in the real world.
%\xinhui{The facial parameters can describe a large-scale 3D space. Each group of the facial parameters can represent one point in the 3D space. Due to the freedom of performance of the identity parameters, a group of identity parameter and another group of facial parameter can represent the same point in same time. As well as some points don't exist in the real world.}
To solve the problem, we apply a facial constraint. Instead of directly predicting identity parameters in the high-dimensional space, a transform is performed to obtain an intrinsic low-dimensional subspace on the identity parameters. 
We collect 20K face images without expression based on Basel Face Model \cite{bfm09}, and obtain a set of identity parameters using our GE-Net with AU parameters all set to zero. 

Suppose $m$ is the mean of identity parameters and $U$ is the projection matrix, which can be obtained by performing singular value decomposition on the covariance matrix of the identity parameters. The formula is defined as:
\begin{equation}
U = SVD((\alpha_{id} - m)^T(\alpha_{id}-m)),
\end{equation}

\begin{equation}
\hat{\alpha_{id}} = (\alpha_{id}-m)U.
\end{equation}  
$\hat{\alpha_{id}}$ means the identity parameters in the low-dimensional subspace. Each of the parameters $\hat{\alpha_{id}}$ can be normalized to [0,1] according to the 20K face images in the subspace. The $\hat{\alpha_{id}}$ is back-projected to the high-dimensional space by multiplying the transpose of the projection matrix. The final predicted identity parameters can be reconstructed as:

\begin{equation}
\label{project_high}
\alpha_{id} = \hat{\alpha_{id}}U^T + m.
\end{equation}

Fig. \ref{generate_images} shows some randomly generated faces without and with the facial constraint.  
Before the input image is fed into the feature extractor, it is aligned to the ``base face" via five facial landmarks. % using affine transformation
The ``base face" is created by rendering a frontal emotionless face image, with the identity parameters set to 0.5.

In the fitting process, we set different learning rates for $(h,\alpha_{exp},\alpha_{id})$. For the identity parameters, the learning rate is $8.0$. For the head pose and AU parameters, we find each parameter has a different impact on the generated facial expression, so we set different learning rates. Specifically, starting from the ``base face", to compute the learning rate for one parameter in $(h,\alpha_{exp})$, we set its value to the maximum value while keeping other parameters zero, and obtain a corresponding segmentation output. We count the different pixels compared with the segmentation output of ``base face", and set the inverse of the ratio of different pixels in the whole image as the learning rate.
%According to the ratio of the change, we initialize the learning rate of $l_r$. 
The optimization can converge well after 30 iterations in the experiment.

%(because the parameters have different changes in the performance of the generated image \tianjia{(what do you mean? dont understand)}. 
%We count the distinction of segmentation result of the generated images, which are generated by maximizing each of the parameters based on the ``base face”. According to the ratio of the distinctions, we initialize the learning rate of $l_r$. \tianjia{(dont understand. what distinction? segmentation result? dont use segmentation anymore. It is just misleading. totally confused.)} The regressive process \tianjia{(where is the regression? you just did optimizaiton.)} of the AU parameters can be regarded as an optimization problem, in which the loss can be minimized via a sequence of the generated images. \tianjia{(dont understand.)}
%The optimization is iterated 30 to 50 times in our experiment.)

\section{Experiments}
%The goal of experiments is to show that our method can achieve  the state-of-the-art performance.

%Among them, we consider the 13 AUs (1, 2, 4, 5, 6, 7, 9, 10, 12, 17, 23, 24, 25), whose occurrence frequency is larger than 10\%.	%displaying various facial expressions of emotion, single AU activation, and multiple AU activation.
%We obtain 329 sequences that are annotated with AUs. 
%13 AUs (1, 2, 4, 5, 6, 7, 9, 10, 12, 17, 23, 25, 26) with occurrence frequency larger than 10\% are considered in our experiments. 
\subsection{Dataset}
{\bf Public Data.} We adopt the Extended Cohn-Kanade database (CK+) \cite{lucey2010extended}, the MMI database \cite{pantic2005web}, FERA2015 \cite{valstar2015fera}, the DISFA \cite{mavadati2013disfa} and the FaceWarehouse \cite{cao2013facewarehouse} to evaluate our method. 
The CK+ database contains 593 sequences from 123 subjects. Approximately 15\% of the sequences are coded by a certified FACS coder, which coincides with the peak frames. 
The MMI database contains more than 2800 samples, which contain static images and image sequences in frontal and in profile view.
The FERA 2015 contains about 140,000 images from 41 subjects. 
Intensities are annotated for 5 AUs (6, 10, 12, 14, 17).
The DISFA database consists of 27 sequences from 27 subjects. Around 130,000 frames are annotated with AU intensity for 12 AUs (1, 2, 4, 5, 6, 9, 12, 15, 17, 20, 25, 26).
%is a spontaneous expression database, which
% FaceWarehouse is a facial expression database, including 2D images and 3D meshes using a Kinect RGBD camera. It contains 150 individuals, for each person, different expressions are captured, including the neutral expression and 19 other expressions such as mouth-opening, smile, kiss, etc. We only use the 2D images to evaluate our method.

\begin{figure*}[ht]
	\begin{center}
		\includegraphics[width=0.9\linewidth]{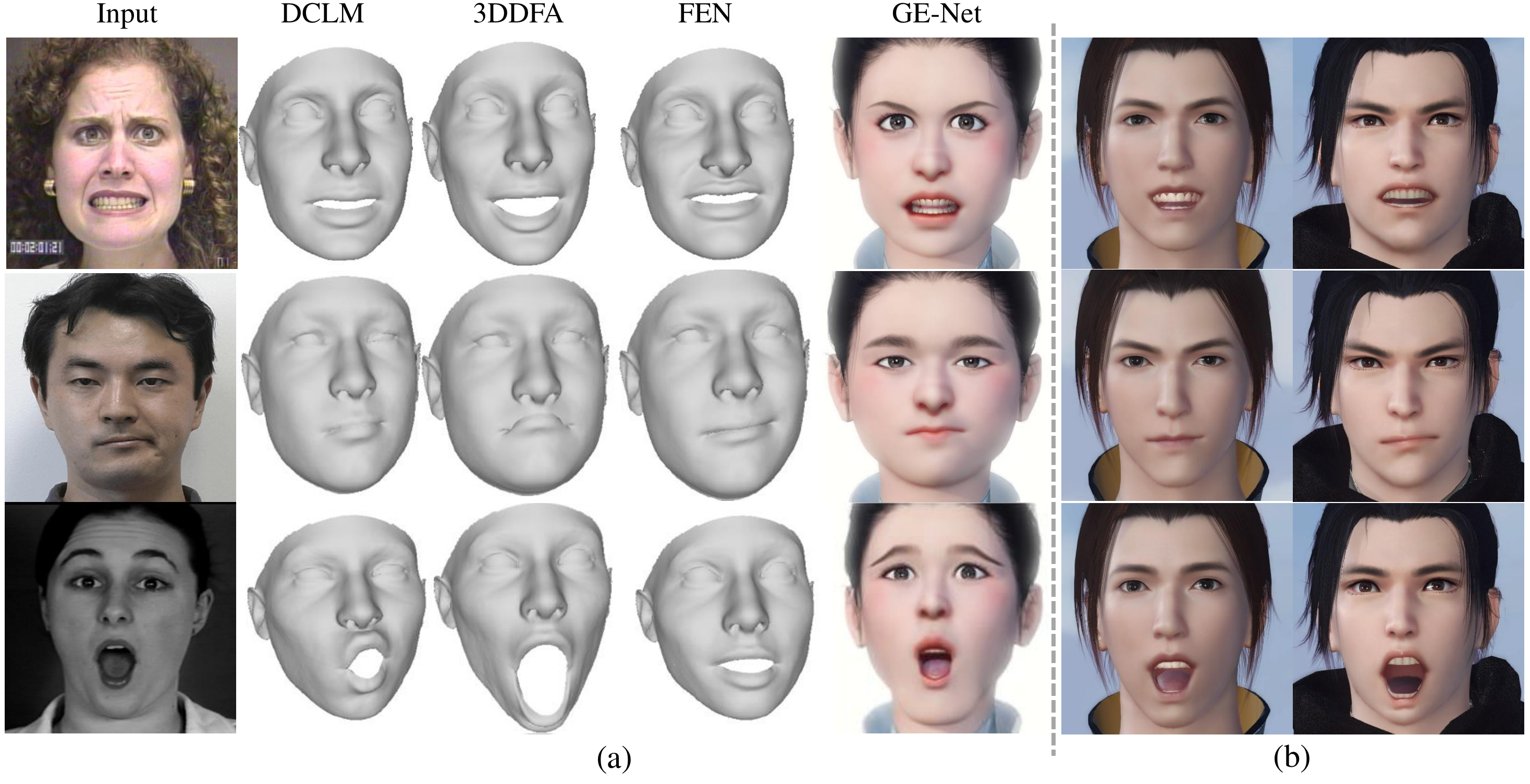}
	\end{center}
	\caption{ Qualitative AU intensity estimation results (a) and the transferred results (b). The results of DCLM \cite{zadeh2016deep}, 3DDFA \cite{zhu2016face} and FEN \cite{chang2019deep}  are from original paper. The results of AU parameters transferred to another 3D model are shown in (b).}
	\label{qulativate_compare}
\end{figure*} 

{\bf Rendered Data.} We render 80,000 face images using Justice Face with random facial parameters. The facial parameters is a 269-dimensional vector with head pose $h\in \mathbb R^{1\times2}$, AU parameters $\alpha_{exp} \in R^{1\times23}$ and identity parameters $\alpha_{id} \in \mathbb R^{1\times244}$.  
specifically, the range of $h$ is from $-28.6^\circ$ to $28.6^\circ$ in pitch and yaw, and we convert the angles to be within $[0,1]$.
We only consider the head pose in pitch and yaw because the image is aligned in the roll, which means simple 2D in-plane rotation. 
The initial values of the head pose should represent a frontal view, so the values are set to 0.5. 
%Unlike the method of F2P \cite{F2p}, because we do not care about the hairstyle, beard style, and lipstick style, the dimensions $d$ of the $\alpha_{id}$ are reduced to 244, 208 of them are continuous values, and the rest are discrete ones. 
The identity parameters consist of 208-dimensional continuous parameters and 36-dimensional discrete parameters.
These discrete parameters are encoded as a one-hot vector and are concatenated with the continuous ones.  
The range of the facial parameters is from 0 to 1. 
The face samples are generated randomly from a uniform distribution of all parameters. 
90\% face samples are used for training the generator $\mathcal{M}$, and the others are used for validating the generator.

%, and the images may be not unnatural
\subsection{Comparison with the state-of-the-art}
{\bf Comparison with weakly and  semi-supervised learning methods.} We compare the proposed method with several state-of-the-art weakly-supervised learning methods (KBSS \cite{zhang2018weakly}, KJRE \cite{zhang2019joint} and CFLF \cite{zhang2019context}) and semi-supervised learning methods (LBA \cite{haeusser2017learning} and 2DC \cite{linh2017deepcoder}). 
LBA \cite{haeusser2017learning} encourages cycle-consistent association chains from the embedding of labeled samples to unlabeled samples. The back is based on the assumption that samples with similar labels have similar latent features. 
2DC \cite{linh2017deepcoder} combines variational auto-encoder and Gaussian Process for AU intensity estimation by joint learning of latent representations and classification of multiple ordinal outputs. 
KBSS \cite{zhang2018weakly} exploits four types of domain knowledge on AUs to train a deep model. 
KJRE \cite{zhang2019joint} jointly learns the representation and estimator with limited annotations by incorporating various types of human knowledge.
CFLF \cite{zhang2019context} uses a learnable task-related context and two types of attention mechanisms to estimate the AU intensity.
Note that KBSS, CFLF and 2DC need labeled data to train their models. Unlike them, we propose a framework for facial AU intensity estimation without annotated AU data.

%we train the deep model by the 3D facial model made by the artist. The 3D facial model can provide synthetic data for our model. 

\begin{figure*}[ht]
	\begin{center}
		\includegraphics[width=\linewidth]{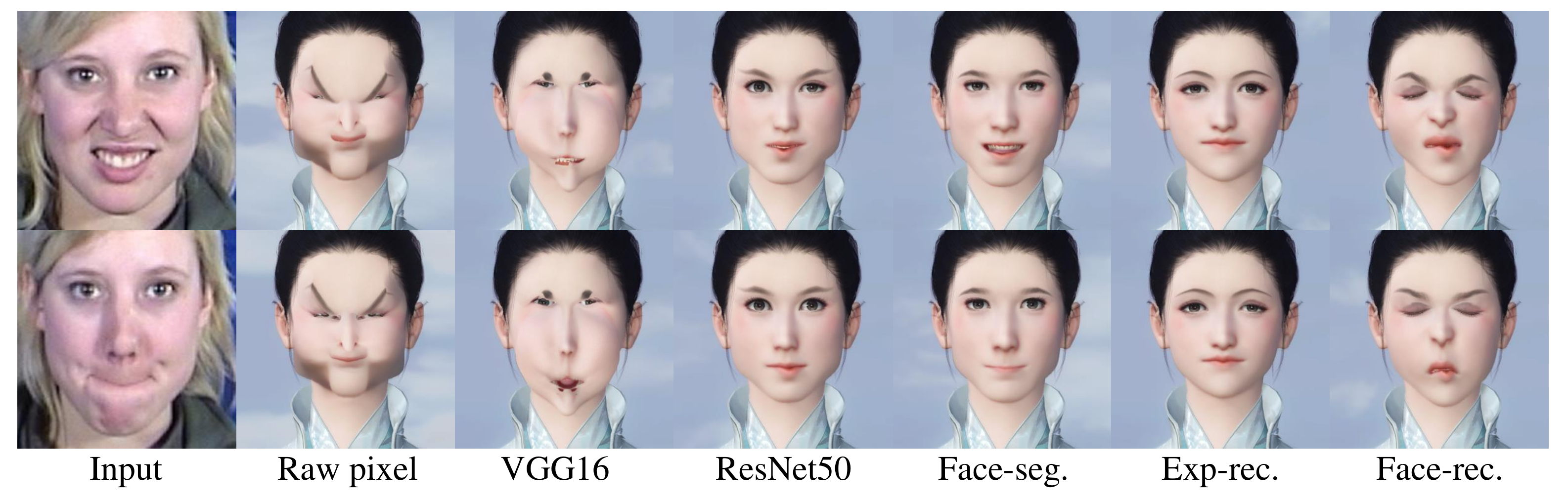}
	\end{center}
	\caption{The results of different feature extractors. The raw pixels and the features of the pre-trained VGG16 are used in the loss, whose results are inferior. The pre-trained ResNet50, the facial expression recognition(``exp-rec.") model ``ResNet" and the face recognition model(``face-rec.") ``EvoLVe" can not distinguish the input images with different expressions. Only the results of the facial segmentation network (``face-seg.\textsf{\textsf{}}") are similar to the real facial images.}
	\label{alation_featextract}
\end{figure*}
\begin{table} 
	\caption{Comparison with the state-of-the-art supervised methods. (*) indicates results taken from the reference.}
	\label{comparison_supervised}
	\begin{center}
		\scalebox{0.9}{
			\begin{tabular}
				{|c| c | c|}
				\hline
				Method &  ICC &	MAE   \\
				\hline
				OpenFace \cite{baltruvsaitis2015cross}& .392 &.421 	 \\
				CNN \cite{gudi2015deep}* &.328 & .423	 \\
				ResNet18 \cite{he2016deep}* &	.270 & 	.483\\
				CNN-IT \cite{walecki2017deep}* &	.377 & .663 \\
				GE-Net & {\bf.641} &{\bf.329}\\
				\hline
		\end{tabular}}
	\end{center}
\end{table}
Intra-Class Correlation (ICC(3,1) \cite{shrout1979intraclass}) and Mean Absolute Error (MAE) are adopted as the metrics for evaluation. The quantitative results are listed in Table \ref{compare_ICC_MAE}, and more quantitative results are listed in the supplementary material. 
We can see that on the FERA2015 dataset, our method achieves superior average performance over other methods on both metrics. 
On DISFA, our method achieves the best average performance on ICC and the second-best on MAE.
Note that ICC and MAE should be jointly considered to evaluate one method. 
LBA tends to predict the intensity to 0, which is the majority of AU intensity. It can achieve good MAE performance, but its ICC is much worse than ours. 
Our method outperforms KBSS and CFLF on DISFA and FERA 2015. During the training phase, these two approaches need multiple frames as input, hence it requires more than one image per face, which is not always available.
%Besides, our method uses the perceptual content to capture the spatial feature.
Besides, the proposed GE-Net utilizes the attention mechanism for the feature extractor, which achieves better results than GENet-O, which does not use the attention mechanism.
It demonstrates the effectiveness of the attention mechanism on the feature extractor.
We also noticed that our method has bad performance on AU 12 in DISFA. It is because Justice Face restricts the expression of lip corner puller, whose movement is smaller than the real face so that the model can express elegant expression.
These results show the superior performance of the proposed method over existing weakly and semi-supervised learning methods.

% 与一些监督的方法比较
{\bf Comparison with supervised learning methods.} We compare our method with several supervised learning methods of AU intensity estimation, including OpenFace \cite{baltruvsaitis2015cross}, ResNet18 \cite{he2016deep}, CCNN-IT \cite{walecki2017deep} and CNN \cite{gudi2015deep}.
CNN \cite{gudi2015deep} is composed of three convolutional layers and a fully connected layer. 
CCNN-IT \cite{walecki2017deep} takes advantage of CNNs and data augmentation. %  based on multiple face datasets
ResNet18 is introduced in \cite{he2016deep}. %a convolutional neural network that
OpenFace \cite{baltruvsaitis2015cross} is based on histograms of oriented gradients and landmark locations.  
These supervised methods require annotating AU intensity of 
each frame in sequences while ours does not need the AU annotations.
The results are shown in Table \ref{comparison_supervised}. Our method achieves better performance on both metrics.

\begin{figure*}[ht]
	\begin{center}
		\includegraphics[width=\linewidth]{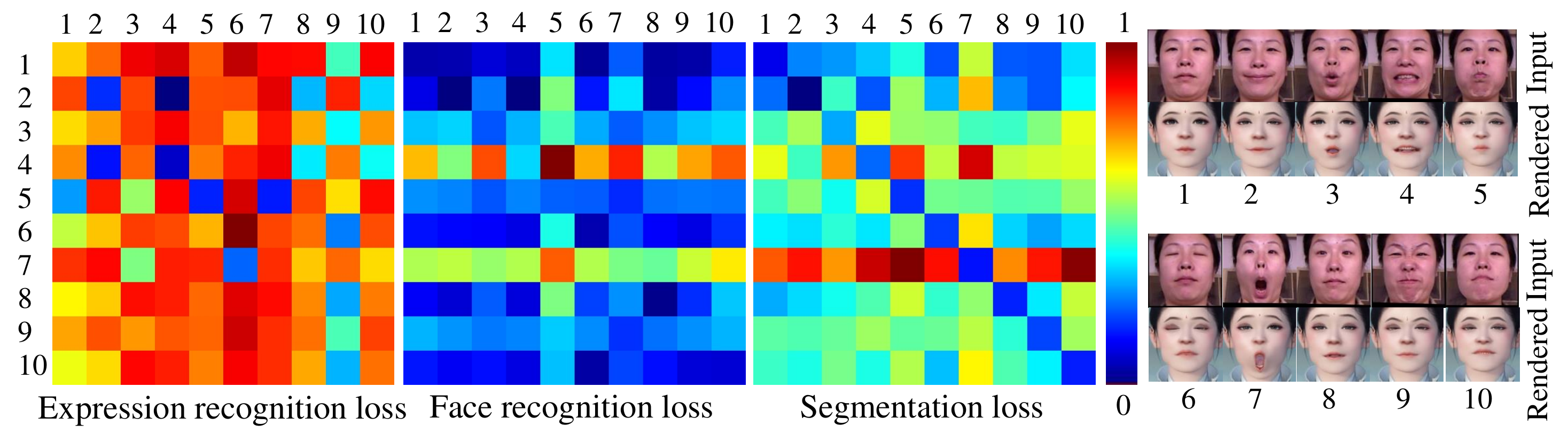}
	\end{center}
	\caption{Expression similarity matrices (plot using color map ``JET") for the losses between the real inputs and the rendered images. The x-axis and y-axis are the inputs and the rendered images, respectively. Ten different expressions are listed on the right. The smaller the value is, the closer is the distance. Only the segmentation loss is low on the diagonal of the matrix, so it can be used to measure the differences.}
	\label{segmentation_valid}
\end{figure*}

\begin{figure*}[ht]
	\begin{center}
		\includegraphics[width=0.9\linewidth]{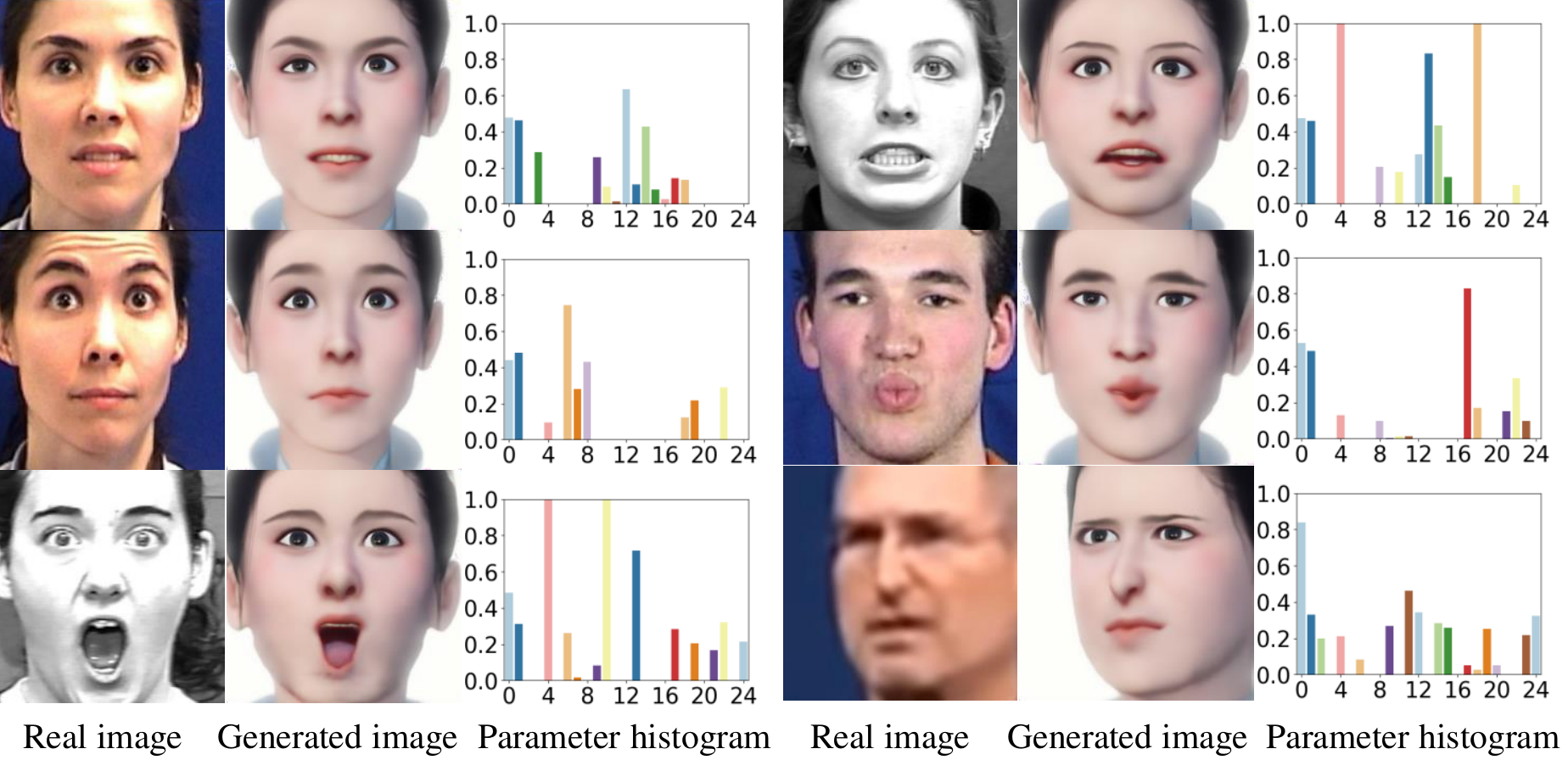}
	\end{center}
	\caption{Some visual results of GE-Net, where the input images, generated images, and the estimated AU parameters are demonstrated. In the parameter histogram, columns 0-1 correspond to the head pose and columns 2-24 correspond to AU parameters. }
	\label{result}
\end{figure*}

%比较我们的方法和一些3dmm合成的方法图像的参数
{\bf Qualitative AU intensity estimation results.} 
%Although the goal of our method is to predict the AU parameters, 
We also show the qualitative results which are rendered using Justice Face with the estimated facial parameters.
Fig. \ref{qulativate_compare}(a) provides several samples of the rendered images of different methods. 
Our method can achieve more accurate estimation than the state-of-the-art methods based on 3DMM. 
As shown in Fig. \ref{qulativate_compare}(b), the AU parameters of the inputs are transferred to another 3D face model.
Fig. \ref{result} shows our additional qualitative results on the mentioned datasets and internet images. (More results are given in the supplementary material.) 
%on MMI, FaceWarehouse, CK+ and DISFA datasets

\subsection{Ablation study}
\label{ablation_sec}
We perform an ablation study to verify the effectiveness of the feature extractor. 
A baseline using the loss in the raw pixel space is explored. 
In addition, instead of using the facial segmentation network, we use the pre-trained ResNet or VGG to extract the features of an image. 
Fig. \ref{alation_featextract} shows the comparison results. 
We can see that the results using the raw pixel loss or the pre-trained VGG16 are rather inferior, with a lot of artifacts. 
The pre-trained ResNet50 can not distinguish the input images with different expressions. In contrast, our facial segmentation network can be aware of the facial shape changes and produce similar expressions as the input images.

%{\bf Why don't we use face or expression recognition model?}
% 增加一个generator的有效性实验 增加了perceptual loss之后 图像清晰了？？？
% 为什么分割是有用的,分割的特征与假人的特征的正相关性
We further validate the effectiveness of the facial segmentation network by comparing it with the facial expression recognition model ``ResNet" \footnote{https://github.com/WuJie1010/Facial-Expression-Recognition.Pytorch} and the face recognition model ``EvoLVe"\footnote{https://github.com/ZhaoJ9014/face.evoLVe.PyTorch} \cite{zhao2018towards}.
A natural idea would be to use a facial expression recognition model to measure the expression similarity between the input photo and the neurally rendered one. 
However, due to the huge domain gap, existing pre-trained models fail to measure such similarity. 
The face recognition model also fails to distinguish the different expressions, which is illustrated in Fig. \ref{alation_featextract}. 
As shown in Fig. \ref{segmentation_valid}, ten different expressions are selected to compute the losses on the FaceWarehouse dataset. We compare the performances of the expression recognition loss, face recognition loss and face segmentation loss using the expression similarity matrix. The expression similarity matrix is normalized and shown in the JET style.
Obviously, a useful metric for expression similarity measurement should have low values on the diagonal of the matrix, while only the segmentation loss meets the requirement well. 

%\footnote{https://github.com/WuJie1010/Facial-Expression-Recognition.Pytorch}
%\footnote{https://github.com/ZhaoJ9014/face.evoLVe.PyTorch}

\section{Conclusion}
In this paper, we propose an unsupervised framework for estimating the head pose, the AU parameters, and the identity parameters from a real facial image without annotated AU data. 
We formulate the method as a differentiable optimization problem by minimizing the differences of the extracted features between the real facial image and the generated one. 
The loss can be back-propagated to the facial parameters because the generator is differentiable. 
Experimental results demonstrate that the proposed method can fit accurate AU parameters and achieve state-of-the-art performance compared with existing methods. 
It can also get accurate AU estimation on internet photos. In the future, we plan to collect a large number of face videos with the various expression of different views. 
The database can be labeled using our GE-Net so that we can exploit faster methods for AU intensity estimation. 
Meanwhile, a more robust method to handle the large change of the head pose is another direction.
%We get the interpretable parameters, which are derived from the underlying facial anatomy. 

{\small
\bibliographystyle{ieee_fullname}
\bibliography{egbib}
}

\onecolumn
\newpage
\appendix
\section{Appendix}
\subsection{More Experiments} 

\begin{figure}[ht]
	\begin{center}
		\includegraphics[width=\linewidth]{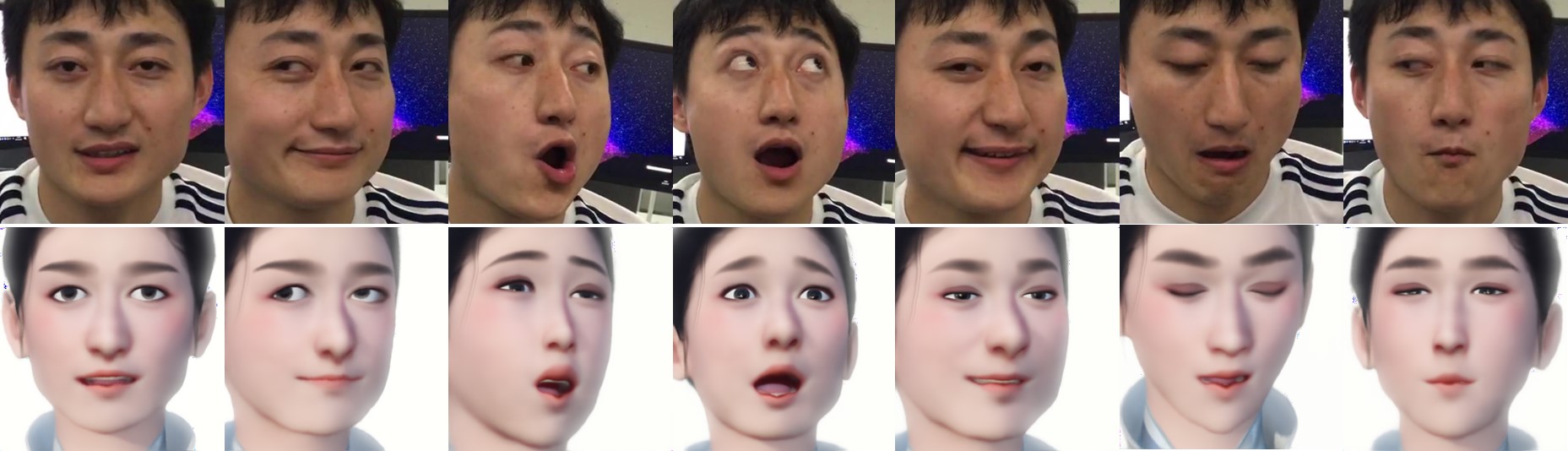}
	\end{center}
	\caption{A video expression transfer example. The upper row shows the several frames of one video, and the bottom row shows the generated images. }
	\label{videoResult}
\end{figure}

{\bf Stability.} 
%The model can convert a sequence of input images to the AU parameters. 
We hope that the predicted AU parameters are stable in the continuous frames of one video. 
The stability is crucial for numerous applications such as expression transfer, facial expression analysis, and human-robot interaction. 
To evaluate the stability, we suppose that the continuous two frames are almost the same in one video. 
So, the stability is evaluated by the differences between the continuous frames, which is calculated by the standard deviation of each AU in the fixed-size sliding windows of continuous frames. 
We compute the standard deviation of ten video sequences from 2 to 10 frames. The ten videos are download from the internet. 
The results of our method are better than OpenFace on different sliding windows. 
The result of 5 frames is shown in Table \ref{std_table}. We also show the results of a video in different frames in Fig. \ref{videoResult}.

\begin{table}[ht]   
	\small
	\caption{The standard deviation on the sliding windows. Bold numbers indicate the best performance.}
	\label{std_table}
	\begin{center} 
		\begin{tabular}
			{p{12mm}<{\centering}p{7mm}<{\centering}p{7mm}<{\centering}p{7mm}<{\centering}p{7mm}<{\centering}p{7mm}<{\centering}p{7mm}<{\centering}p{6mm}<{\centering}p{7mm}<{\centering}p{7mm}<{\centering}p{7mm}<{\centering}p{7mm}<{\centering}p{7mm}<{\centering}p{7mm}<{\centering}}
			\toprule
			AU & 01 & 02 & 03 & 04 & 05& 06 & 07 & 09 &10&12 &14& 15 &15-2\\
			\midrule 
			OpenFace & 0.247 & 0.261 & -- & 0.114 & 	0.267 &0.139 & 0.248 & \textbf{0.027} & 0.178 &0.144 & 0.218 & --&0.055 \\
			
			Ours & \textbf{0.032} &\textbf{0.013} &0.034 &  \textbf{0.045} & \textbf{0.096} &\textbf{0.059} & \textbf{0.055} &0.034 &\textbf{0.038} &\textbf{0.047} & \textbf{0.044}& 0.064 & \textbf{0.034}  \\
			\hline
			AU & 17 & 18& 20 &  23& 25 & 27&  28&29&34& 45& 45-2&47 & 48  	 \\
			\hline
			OpenFace & \textbf{0.017} & --&0.162 &0.077 & 0.226 &-- & -- &-- & --& 0.221 & --& --&--\\
			Ours & 0.042 &0.054 & \textbf{0.091} & -- & \textbf{0.073} & 0.067 &0.071& 0.053& 0.050 &\textbf{0.102}  & 0.124& 0.024 &0.060\\
			\hline
			%\bottomrule %添加表格底部粗线
			
		\end{tabular}
	\end{center}
\end{table}

\begin{figure}[ht]
	\begin{center}
		\includegraphics[width=0.85\linewidth]{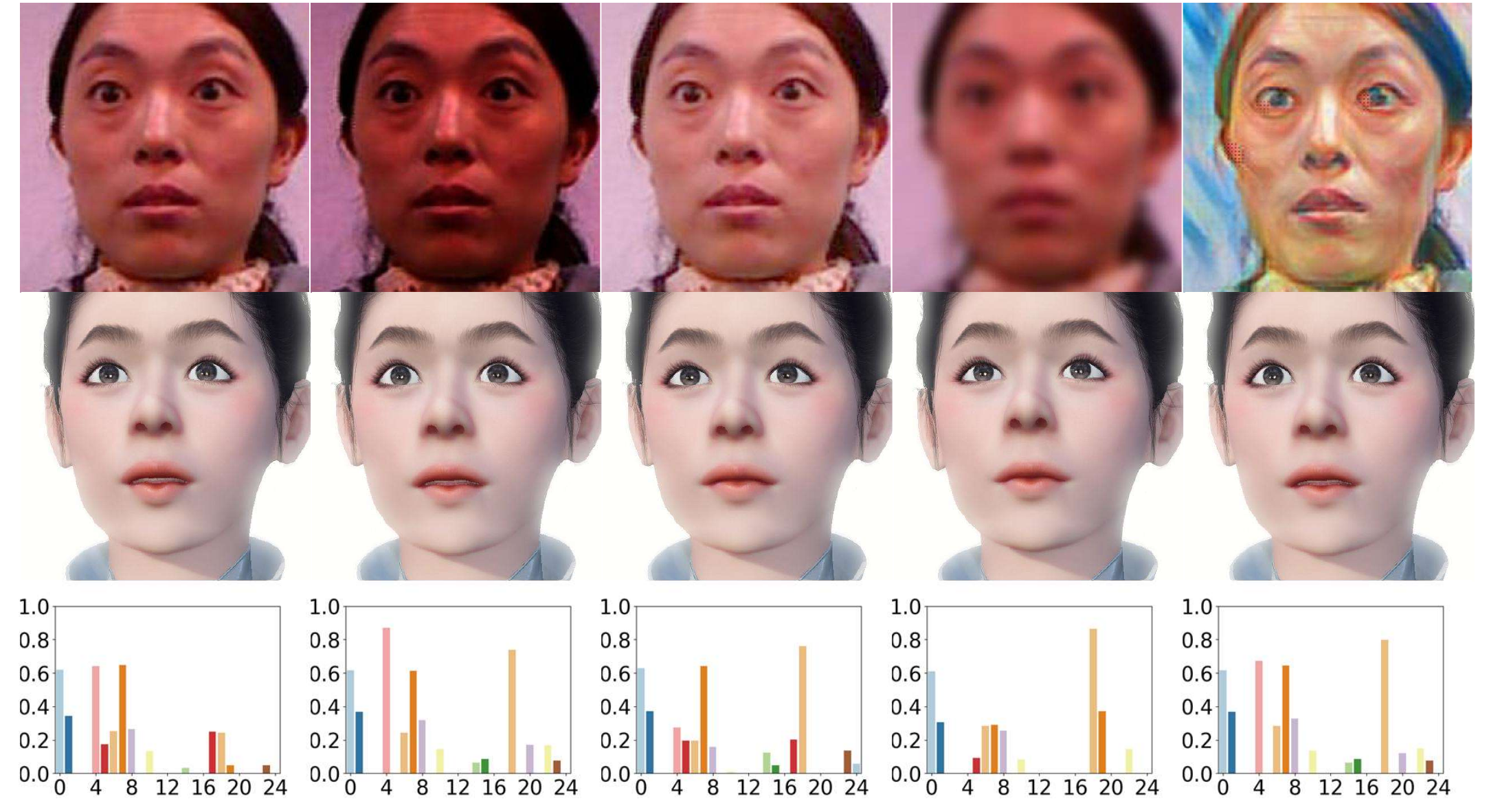}
	\end{center}
	\caption{Robustness test. The network is robust to change in lighting, resolution, and style. From top to bottom, we show the five input images, the rendered images, and the corresponding AU parameter histograms.}
	\label{robustResult}
\end{figure}

{\bf Robustness to Varying Conditions.}
Our network uses facial segmentation features as the input of the loss, yielding results robust to changes in lighting, resolution, and style.
To verify it, we add robustness experiments in FaceWarehouse dataset \cite{cao2013facewarehouse}. Different lights, gaussian blur, and style transfer \cite{johnson2016perceptual} are used to the dataset.
Some results are shown in Fig. \ref{robustResult}. We qualitatively demonstrate this robustness by varying conditions for a single subject, and the results show consistent output. 
The first row is the input image. The rendered images and the corresponding histograms of the facial AU parameters are shown in the second and the third row.
We also noticed that our method has bad performance on the upper lid raiser (column 4 in the histogram). 
It is because Justice Face restricts the expression of the upper lid raiser, whose movement is smaller. (please refer to Fig. \ref{au_parameters}.) 

In our paper, the useful action unit and the corresponding name are listed in Table \ref{au_table} based by row and then column.
Each of AU parameters is set to the max value, and the corresponding rendered image to base face is listed in Fig. \ref{au_parameters}.

\begin{table*} [ht]
	\caption{The AUs and corresponding name.}
	\label{au_table}
	\begin{center}
		\setlength{\tabcolsep}{3.5mm}{
			\begin{tabular}
				{c|c||c|c||c|c}
				\hline
				Name & AU & Name & AU & Name & AU    \\
				\hline 
				\hline
				Left Eye Blink & 45 & Right Eye Blink & 45-2 &Upper Lid Raiser & 5 \\
				\hline
				Cheek Raiser & 6 & Inner Brow Raiser & 1 & Left Outer Brow Raiser & 2 \\
				\hline
				Right Outer Brow Raiser & 3 & Brow Lowerer & 4 &  Mouth Stretch & 27\\
				\hline
				Nose Wrinkler  & 9 & Upper Lip Raiser & 10 & Down Lip Down & 25 \\
				\hline
				
				Lip Corner Puller & 12 & left mouth press & 14 & right mouth press & 14-2 \\
				
				\hline
				
				Lip Pucker  &18 & Lip Stretcher & 20 & Lip Upper Close & 24 \\
				\hline
				
				Lip Lower Close & 17 & Cheek Puff &  34 & Lip Corner Depressor  & 15\\
				\hline
				
				Jaw Left & 47 &  Jaw Right & 48  \\	
				\hline%添加表格底部粗线
				
		\end{tabular}}
	\end{center}
\end{table*}

{\large \bf Compared with OpenFace.} We provide a comparative experiment to investigate the effectiveness of our GE-Net. The results are shown in Table \ref{compare_table}. The GE-Net achieves better performance in MAE and ICC on average and many AUs, which demonstrates the effectiveness of the attention mechanism.
\begin{table*}
	\caption{The MAE on CK+ and MMI. Bold numbers indicate the best performance. GENet-O means our method without attention.} 
	\label{compare_table}
	\begin{center} 
		\begin{tabular}{|p{1mm}<{\centering}|p{12mm}<{\centering}|p{17mm}<{\centering}|p{4mm}<{\centering}p{4mm}<{\centering}p{4mm}<{\centering}p{4mm}<{\centering}p{4mm}<{\centering}p{5mm}<{\centering}p{4mm}<{\centering}p{4mm}<{\centering}p{4mm}<{\centering}p{4mm}<{\centering}p{4mm}<{\centering}p{4mm}<{\centering}p{4mm}<{\centering}p{5mm}<{\centering}|} %12 au 
			\hline
			~& Database &AU & 01 & 02& 04 &05& 06& 07& 09& 10& 12& 17& 23 &24 &25 & Avg \\
			\cline{2-17}
			
			\multirow{7}{*}{\rotatebox{90}{MAE}}&\multirow{3}{*}{CK+} & OPENFACE & .45 & .41 & .74 & .33 & .65 & 1.34 & .58 & .43 &{\bf.46} & .94 &.56 &.60& .63 &  .624  \\
			
			%	~& ~ & GENet-O & .41 & .37 & .69 & .28 & .33 & .31 & .54 & .14 & .68 & .87 & .49 &  - & .39 & .470\\
			~&~ & GE-NET & {\bf.40} & {\bf.27} & {\bf.66} & {\bf.27}&  {\bf.30} & {\bf.30} &
			{\bf.51} & {\bf.14} &  .68 & {\bf .77} & {\bf.48} & - &
			{\bf.39} &  {\bf.443} \\
			\cline{2-17}

			&~& AU & 01 & 02& 04 &05& 06& 07& 09& 10& 12& 17& 23 &25 &26 & Avg \\
			\cline{3-17}
			~&\multirow{3}{*}{MMI} & OPENFACE & 1.09 & .86 & {\bf.23} & .68 & .12 & .21 & {\bf.16} & .27 &{\bf .20} &.60 & {\bf.27} & .46 & .32  & .421  \\
			
			%	~&~ & GENet-O & .20 & .25 & .36 & .64 & .11 & {\bf.04} &  .21 & .46 & .49 & .38 & .36 & .37 &.11 & .321 \\ 
			
			~& ~ & GE-Net &{\bf.12} & {\bf.18} & .23 & {\bf.09} & {\bf.11} & .11 &
			.17 & {\bf.33} &  .56 &{\bf.24} & .324&{\bf.32} & {\bf.10} & {\bf.231} \\
			\hline 
			
			% 下面是ICC的值
			~& ~ &AU & 01 & 02& 04 &05& 06& 07& 09& 10& 12& 17& 23 &24 &25 & Avg \\
			\cline{3-17}
			
			\multirow{7}{*}{\rotatebox{90}{ICC}}&\multirow{3}{*}{CK+} & OPENFACE & .36 & .48 & .24 & .51 & .29 & .35 & .51 & .34 &{\bf.58} & .53 &.48 &.31 & .61 &  .430  \\
			
			%	~& ~ & GENet-O & .41 & .37 & .69 & .28 & .33 & .31 & .54 & .14 & .68 & .87 & .49 &  - & .39 & .470\\
			~&~ & GE-NET & {\bf.58} & {\bf.53} & {\bf.46} & {\bf.78}&  {\bf.70} & {\bf.71} &
			{\bf.80} & {\bf.36} &  .42 & {\bf .75} & {\bf.61} & - &
			{\bf.63} &  {\bf.611} \\
			\cline{2-17}  
			
			&~& AU & 01 & 02& 04 &05& 06& 07& 09& 10& 12& 17& 23 &25 &26 & Avg \\
			\cline{3-17}
			~&\multirow{3}{*}{MMI} & OPENFACE & .50 & .31 & .24 & .44 & .43 & .47 & .31 & .45 &{\bf .38} &.27 & {\bf.48} & .46 & .32  & .389  \\
			
			%~&~ & GENet-O & .20 & .25 & .36 & .64 & .11 & {\bf.04} & .21 & .46 & .49 & .38 & .36 & .37 &.11 & .321 \\ 
			
			~& ~ & GE-Net &{\bf.53} & {\bf.48} & {\bf.38} & {\bf.63} & {\bf.47} & {\bf.53} &
			{\bf.57} & {\bf.49} &  .36 &{\bf.48} & .41 &{\bf.60} & {\bf.51} & {\bf.495} \\
			\hline 
			
		\end{tabular}
	\end{center}
\end{table*}

\newpage
\section{AU parameters expression on the 3d facial model}
\begin{figure}[H]
	\begin{center}
		\includegraphics[width=0.98\linewidth]{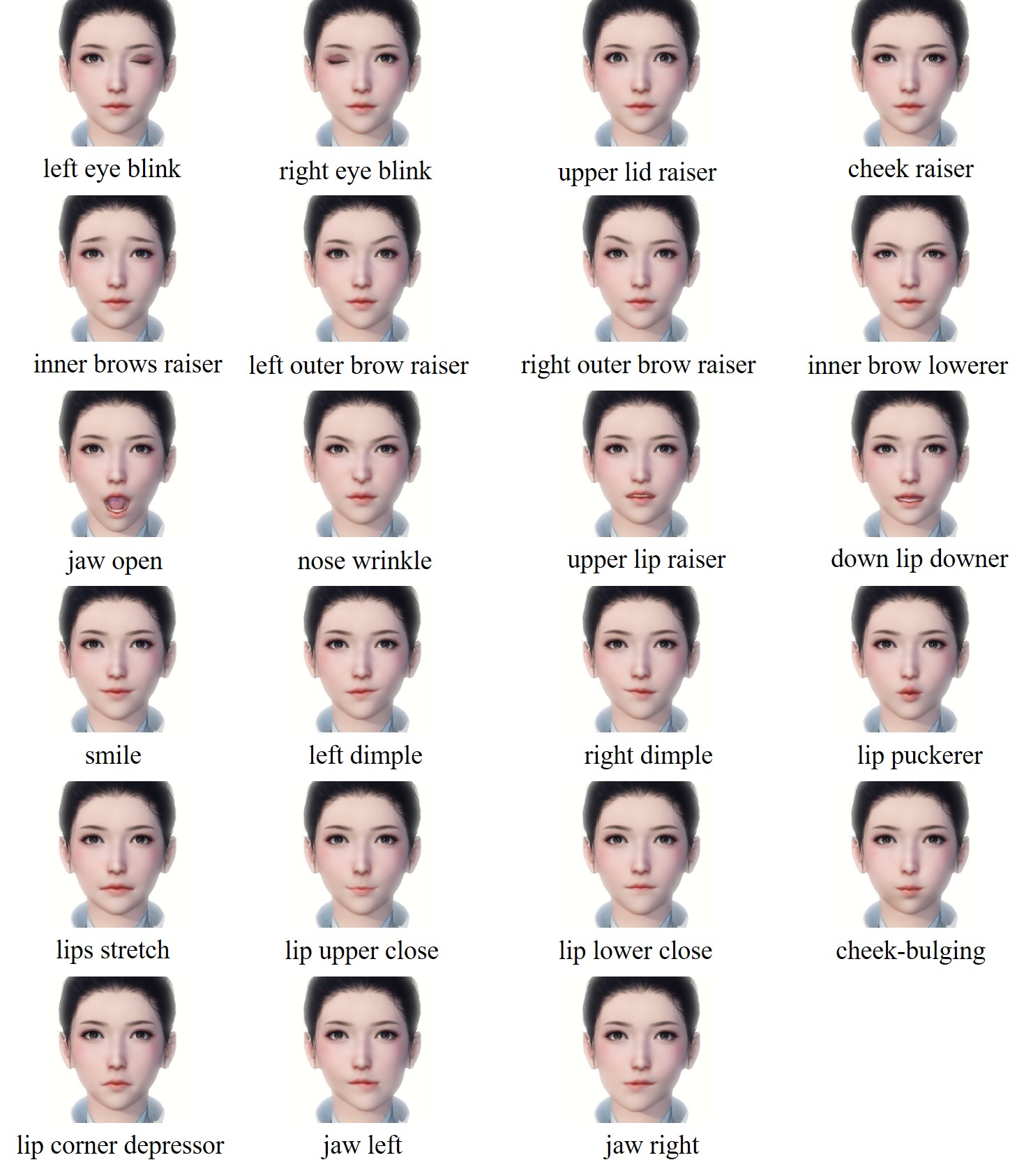}
	\end{center}
	\caption{The generated images with the corresponding AU parameters.}
	\label{au_parameters}
\end{figure}

There are a little bit differences between the AU parameters from the 3d facial model and the AUs from the FACS system (\emph{e.g}. we separate the dimple to left dimple and right dimple), as shown in Fig. \ref{au_parameters}.

\newpage
\section{Training samples of our generator}
\begin{figure}[H]
	\begin{center}
		\includegraphics[width=0.98\linewidth]{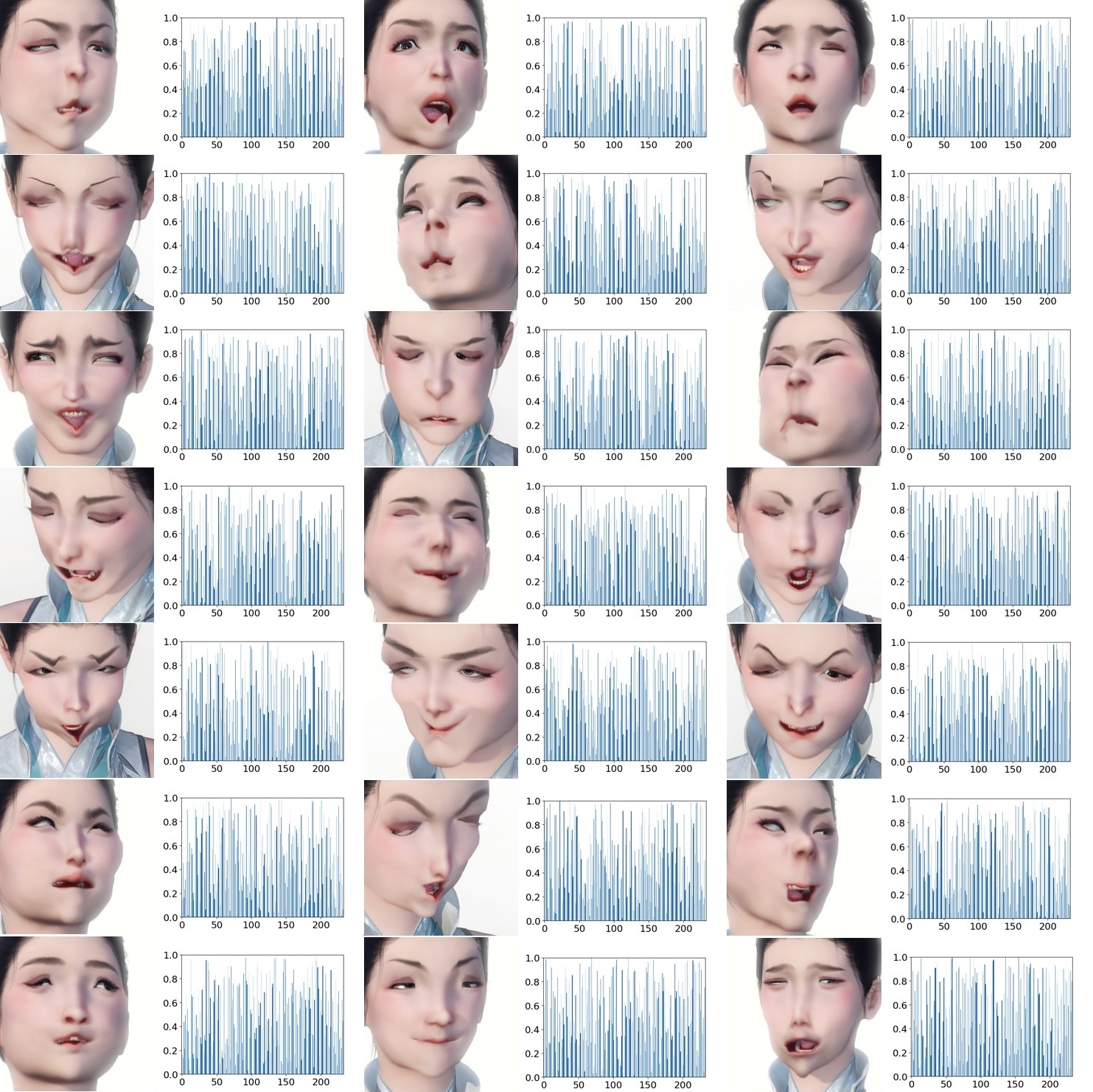}
	\end{center}
	\caption{Training samples of our generator and the corresponding parameters.}
	\label{generrator_data}
\end{figure}

In the training process, we train our generator with randomly generated game faces, as shown in Fig. \ref{generrator_data}, which are decided by the head pose, AU parameters and identity parameters.
% 对我们游戏中的维度进行描述

\section{More examples of our method}
\begin{figure}[H]
	\begin{center}
		\includegraphics[width=\linewidth]{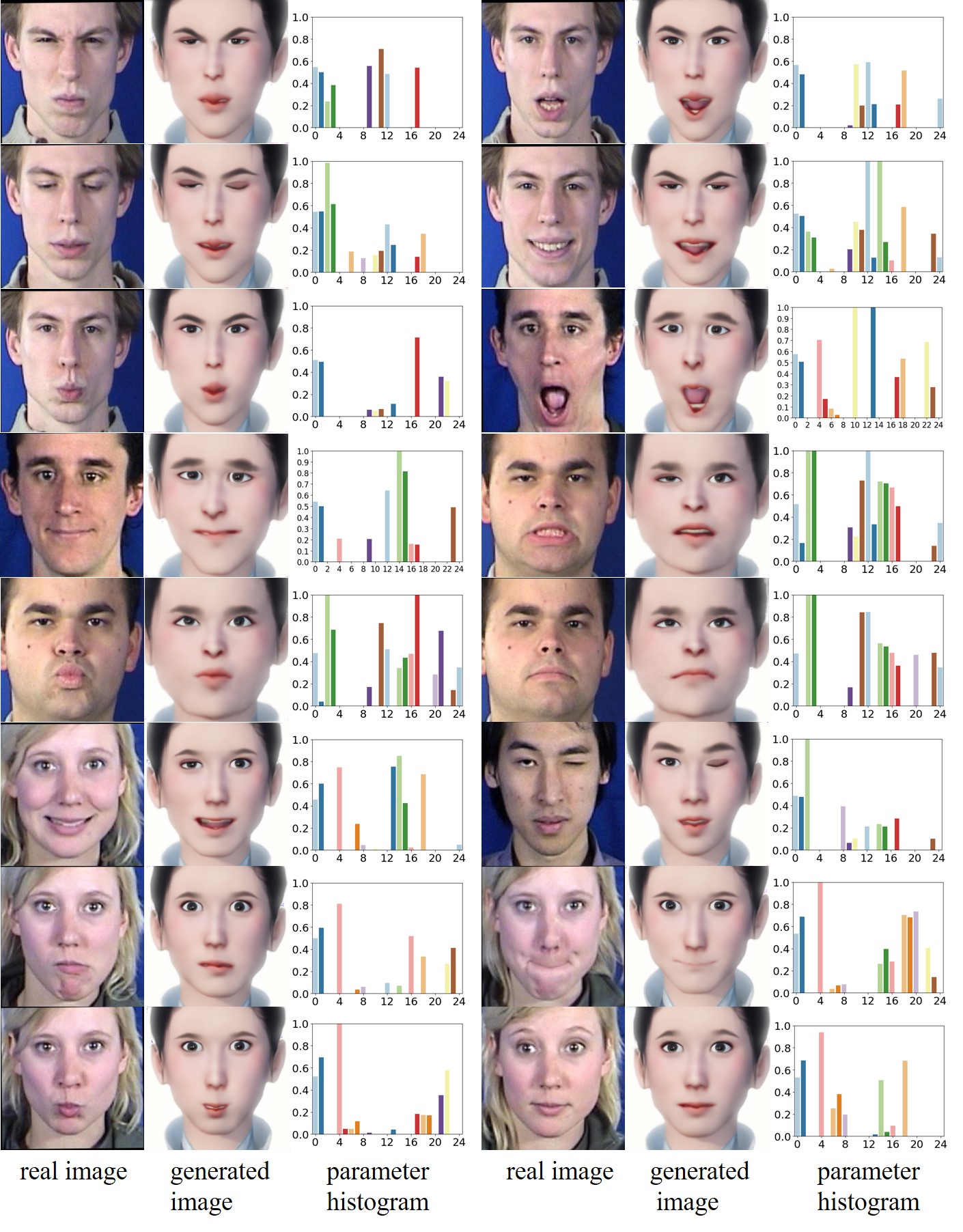}
	\end{center}
	\caption{More results of our method on MMI.}
	\label{result_1}
\end{figure}

\begin{figure}
	\begin{center}
		\includegraphics[width=\linewidth]{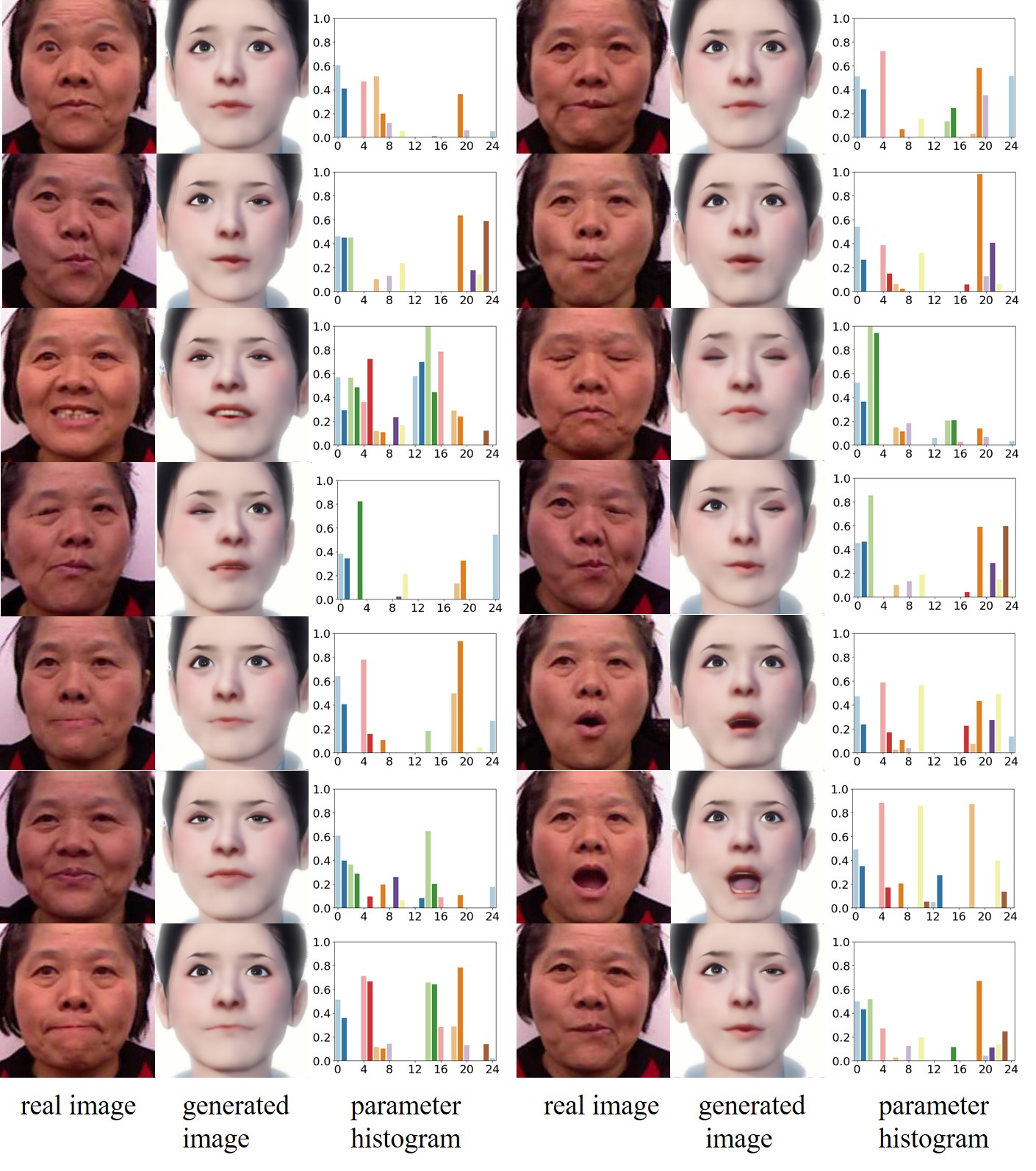}
	\end{center}
	\caption{More results of our method on FaceWarehouse.}
	\label{result_2}
\end{figure}

More examples of our method are shown in Fig. \ref{result_1} and \ref{result_2}. Due to richer expressions on FaceWarehouse dataset, we show more examples.

\end{document}